\documentclass[11pt]{article}

\usepackage[utf8]{inputenc}
\usepackage{amsmath, amssymb, amsthm} 
\usepackage{graphicx}                  
\usepackage{xcolor}                     
\usepackage{hyperref}                   
\usepackage{geometry}                   
\usepackage{booktabs}

\title{Transformer-based Models to Deal with Heterogeneous Environments in Human Activity Recognition}

\author{
Sannara EK, François PORTET, Philippe LALANDA \\
Univ. Grenoble Alpes, CNRS, Grenoble INP, LIG F-38000, Grenoble, France \\
\texttt{\{sannara.ek, francois.portet, philippe.lalanda\}@univ-grenoble-alpes.fr}
}

\date{} 

\begin{document}

\maketitle

\begin{abstract}
Human Activity Recognition (HAR) on mobile devices has been demonstrated to be possible using neural models trained on data collected from the device's inertial measurement units. These models have used Convolutional Neural Networks (CNNs), Long Short-Term Memory (LSTMs), Transformers or a combination of these to achieve state-of-the-art results with real-time performance. However, these approaches have not been extensively evaluated in real-world situations where the input data may be different from the training data. This paper highlights the issue of data heterogeneity in machine learning applications and how it can hinder their deployment in pervasive settings. To address this problem, we propose and publicly release the code of two sensor-wise Transformer architectures called HART and MobileHART for Human Activity Recognition Transformer. Our experiments on several publicly available datasets show that these HART architectures outperform previous architectures with fewer floating point operations and parameters than conventional Transformers. The results also show they are more robust to changes in mobile position or device brand and hence better suited for the heterogeneous environments encountered in real-life settings. Finally, the source code has been made publicly available.
\end{abstract}

\textbf{Keywords:} Human Activity Recognition, Machine Learning, Transformers, Data Heterogeneity

\section{Introduction}


Pervasive computing promotes the integration of smart electronic devices in our living and working spaces to provide a wide variety of services. This concept defined several years ago \cite{weiser1991computer,becker}, is now implemented in many fields and meets an undeniable success beyond the academic spheres. This is due to many factors, including the diversity and quality of the sensors now available, but also the development of new algorithms able to process the information provided by these sensors. In this respect, the use of machine learning (ML) techniques marked a breakthrough. They initiated the development of more advanced services, previously unreachable because of the lack of algorithmic tools to cope with the complexity of pervasive environments. In short, ML algorithms make use of massive data sets collected in the field to automatically train models that can make relevant predictions with new, unseen data. 

The field of wearable devices is particularly representative of the features mentioned here before. Indeed, devices like smartphones or smartwatches are today equipped with high-quality Inertial Measuring Units (IMU) sensors, like accelerometers and gyroscopes. These sensors allow many new applications for health, wellness, or simply informational services \cite{10.1007/978-3-030-96068-1_1}. Wearable devices are in particular used to identify and monitor basic activities automatically. This feature, usually referred to as Human Activity Recognition (HAR), includes recognition of movements such as walking or jumping and postures like standing or sitting. Such detection is the basic building block for identifying more complex activities, often requiring introducing a broader context of execution \cite{roggen2012opportunistic}.


If many approaches have addressed HAR using probabilistic models \cite{blachon:hal-01082580}, most current solutions for HAR are based on deep learning techniques using continuous streams of data provided by the worn sensors. Many studies have employed Convolutional Neural Networks (CNN) \cite{lecun1995convolutional} to learn kernels to extract representative features from the input data, Long Short-Term Memory (LSTM) \cite{gers2000learning} to allows sequential processing of the data, Transformers \cite{IFConvTransformer2022,Kim2022} to incorporate attention mechanisms on the inputs or a combination of these to achieve state-of-the-art results \cite{gu2021survey,challa2021multibranch}. Those approaches have demonstrated high recognition rates reaching 90\% or more, but essentially on relatively small and often quite homogeneous datasets (\textit{i.e.} constituted by a small number of people, or with the same sensors worn in the same way). However, these studies do not provide details on their computation and memory costs. Furthermore, performances stay difficult to compare since experimental training and testing data are not standardized. Finally, even recent Transformer-based architectures \cite{IFConvTransformer2022,Kim2022} did not publicly release their code, making reproducibility and comparison with state-of-the-art challenging. 

Despite the reported impressive accuracy of these models, they are actually very difficult to deploy in the real world. Indeed, most models trained on a specific dataset, cannot be applied in realworld settings because of the mismatch between the training data and the one the model will have to deal with. This is particularly acute in pervasive applications. In practice, this is due both to statistical heterogeneity (differences in usage and environment) and system heterogeneity (differences in system traits) \cite{li2020federated}. 

The objective of this paper is twofold\footnote{Note that a pre-print version of this paper is available at \url{https://arxiv.org/abs/2209.11750}}. First, it is intended to demonstrate that the problem of heterogeneity is critical and hinders the broad deployment of many ML-based applications in pervasive settings. To do so, we have set up an experimentation environment based on several HAR datasets presenting a variety of situations (different users, different devices, and different ways to wear them). We then implemented a standard CNN model, trained it on different configurations, and tested them in unseen situations. This first study constitutes our baseline. Then, we present a sensor-wise transformer architecture adapted for the IMUs-based HAR domain called HART, standing for Human Activity Recognition Transformer. We also propose an extension called MobileHART which adds layers to better capture temporal dependencies in the sensor data. These architectures have been extensively tested on the experimentation environment previously mentioned. Results showed that they achieve better performance than other state-of-the-art models in homogeneous situations and that they are also more robust to data heterogeneity encountered in real life, such as unseen mobile positions or device brands. To make the reproducibility of results possible, all code and data partitions have been made publicly available at: \url{https://github.com/getalp/Lightweight-Transformer-Models-For-HAR-on-Mobile-Devices}.

The paper is organized as follows. First, the problem tackled here is clearly stated, and our experimentation environment is described. Then our proposed transformer-based model is detailed in Section 3. Section 4 presents the evaluation results of this model and is followed by an analysis in section 5. Finally, the article concludes with a discussion of the results obtained and a projection on future work.

\section{Problem statement and baseline experiments}

\subsection{Problem statement}


The problem we seek to characterize and solve is the development of models built with (limited) data from specific environments that have to be used in more varied environments. This problem can be termed as robustness to data heterogeneity is frequent in pervasive computing, where execution environments and users are often diverse and non-deterministic. In many cases, all types of environments and users cannot be equipped to collect and annotate the data necessary to train generic models. This would be too costly and require a measurement campaign out of reach for most companies. In this paper, we also argue that standard lightweight model architectures, like CNNs or LSTMs, cannot deal well with this heterogeneity problem.


On this subject, the example of HAR is perfectly illustrative. This field is confronted with two types of heterogeneity: the one related to users (with different ages, fitness levels, habits, ways to carry their phones, etc.) and the one related to devices (different smartphone brands, versions, sensor qualities, etc.). It is difficult to collect data for all situations and all user profiles. Therefore, it is necessary to develop models capable of transferring what has been learned from training data to execution contexts that differ from those used at training time. We will see that this is not the case today. That is, a model trained using data from such smartphones will not be effective for a smartphone outside this scope. Similarly, a model trained with certain positions will not be used satisfactorily by a person carrying his smartphone differently.

\subsection{Experimentation datasets}\label{sec:datasets}


To cover various situations, we have selected 5 public HAR datasets. These datasets provide data on different users, different types of smartphones, and different ways of wearing them. We can then train models on certain conditions and test them on others. 



The first dataset is UCI \cite{Anguita2013APD} which was collected with a Samsung Galaxy~S~II, with a sampling rate of 50 Hz. It was positioned on the subject's waist. Experiments were conducted on 30 participants with an age range of 19-48 years old. The dataset was collected in a controlled indoor lab environment using artificial situations. 


The MotionSense dataset \cite{Malekzadeh:2018:PSD:3195258.3195260} had 24 subjects, with a variety of ranges regarding gender, age, weight, and height, performing long and short trials. The test was conducted in the same environment for all subjects using an Apple iPhone~6s, with a sampling rate of 50 Hz, that was kept in the subject's front pocket \cite{tang2020exploring,chen2021deep}.

The HHAR dataset \cite{10.1145/2809695.2809718} is based on 9 participants. Each carried 8 Android smartphones of 4 different types within a compact pouch and 4 Android smartwatches on the arm. All 12 devices recorded the activities using the on-device accelerometer and gyroscope measurements to their respective maximum sampling rate (between 50 Hz and 200 Hz). 
 
The RealWorld (RW) dataset \cite{realword} is based on 15 subjects using 7 different devices with a sampling rate of 50 Hz. It is one of the largest available datasets of this kind. The dataset was recorded from different body positions (chest, forearm, head, shin, thigh, upper arm, and waist) and presented a statistically heterogeneous learning environment.  
 
The Sussex-Huawei Locomotion (SHL) Preview dataset {\cite{8418369}} was recorded by 3 different users with 4 different devices and body positions. We only considered the accelerometer and gyroscope data for our experiment. Data was collected using Huawei Mate~9 smartphones with a sampling rate of 100 Hz. The dataset is large compared to previous datasets and presents challenging classification tasks on locomotion-based activities. 


\begin{table}[!bht]
\centering
\caption{Datasets properties.}

\begin{tabular}{crc}
\hline
Dataset   & Data Samples & Activities \\ 
\hline
UCI & 10,299  & 6 (ST,SD,W,U,D,L) \\ 
MotionSense & 17,231  & 6 (ST,SD,W,U,D,R) \\ 
HHAR & 85,567 & 6 (ST,SD,W,U,D,BK)\\
RealWorld & 356,427 & 8 (ST,SD,W,U,D,J,L,R)  \\
SHL Preview & 640,144 & 8 (ST,W,R,BK,C,BS,T,SW)  \\
\hline
Combined & 1,109,668 & 13 Unique Activities\\
\hline
\end{tabular}
\label{fig:datasetProp}
\end{table}

Since datasets were recorded with different sampling rates, we down-sampled all of them to 50 Hz after having applied an Anti-Alias filter. A recent survey regarding HAR on smartphones \cite{overviewHar} has shown that the optimal sampling frequency is between 20 Hz and 50 Hz and that accelerometers and gyroscopes are the most adequate sensors for classification. 

After down-sampling, we only used raw signals that we normalized using channel-wise z-normalization. We used a window-frame size of 128 (2.56s) with an overlap of 50\% over the 6 channels from the accelerometers and gyroscopes. UCI is the only dataset that came pre-partitioned by train and test sets. Thus we used these official partitions. For the four other datasets, we partitioned them following a 70\% train-set, 10\% development-set, and 20\% test-set partition with no data overlap. 

\subsection{Evaluation of robustness with Standard CNN}\label{sec:evaluation}

We have then evaluated to which extent state-of-the-art models are robust to data heterogeneity. To do so, we implemented a 2-layered CNN model, and a CNN-LSTM (DeepConvLSTM) model \cite{Ignatov2018RealtimeHA,ordonez2016deep} that are presented as state-of-the-art models within the HAR community (see performances in Table~\ref{tab:fourResult}). To show the vulnerability of the CNN model to data heterogenity
, we performed a leave-one-out position evaluation with the RealWorld dataset. Here, 1 position out of 7 was left out as a test set, and training was performed on the remaining 6 positions. This process was repeated for all other positions to best evaluate the model's robustness against statistical heterogeneity. 
We also performed a leave-one-out device types evaluation with the HHAR dataset and followed the same process with the device types instead of the positions. This enabled us to evaluate the models on their robustness against system heterogeneity. Results are shown in Table \ref{fig:CNNEval}. 


        

\begin{table}
  \caption{CNN model performances with Realworld and HHAR dataset and with unseen positions (left part) and unseen devices (right part)}
   \begin{center}


  \begin{tabular}{cc|cc}
    \hline
  \multicolumn{2}{c}{RealWorld}&  \multicolumn{2}{c}{HHAR}\\
  \hline
  Position & F-Score & Device & F-Score \\
  \hline
  All & 92.62 & All & 96.91 \\
  \hline
  Chest &  \textbf{67.09} & S.Gear & 65.58 \\
  Forearm & 37.19 & LG.Watch & 59.88 \\
  Head & 47.80 & Nexus4 & \textbf{96.33} \\
  Shin & 45.50 & S3 & 85.70\\
  Thigh & 51.76 & S3 Mini & 91.46 \\
  Upperarm & 56.83 & S+ & 78.28\\
  Waist & 45.78 &  &  \\
    \hline
  \end{tabular}
   \end{center}
   \label{fig:CNNEval}
\end{table}

The results show an F-Score for CNN of more than 92\% for RealWorld and 96\% for HHAR (the `all' row in Table~\ref{fig:CNNEval}) when train and test partitions are made with the same distribution. This hides the fact that such a model will not be able to deal with mismatches in train and test data. For instance, evaluation in positions that were not in the training data is challenging for the CNN model. In this evaluation, the best F-Score obtained was 67.09\% on the chest position. The most challenging case, with an F-score of 37.19\%, is when inferring solely on the forearm. Let us note, however, that in this case, the inferred position data were collected using a smartwatch, which amounts to adding a second form of heterogeneity.

Results on the HHAR dataset hint that the models can better generalize on specific unseen devices, as exemplified by a strong F-score of 96.33\% on the Nexus4. The device with the lowest score was the LG.Watch with an F-score of 59.88\% followed by an F-score of 65.58\% on the S.Gear. However, since the LG.Watch, and S.Gear are smartwatches, the negative results are likely due to differences in position (i.e. watches are mostly on the arm while smartphones are mostly on the hips). 




Those results show that the domain of HAR is highly challenging as soon as we consider the need to generalize on different positions and devices (which, in fact, correspond to realistic situations). The findings here show that statistical heterogeneity plays a more significant adversarial role than system heterogeneity, and the effects are drastically worsened when both statistical and system heterogeneity is present. 
The reported outcome is in contrast to past studies \cite{Ignatov2018RealtimeHA,s18113726,challa2021multibranch} in the wearable HAR domain. Previous results, trained and tested on the same device, orientation, and on-body position, have indicated only the need for lite and conventional network architectures for good-performing models. However, it appears that the most frequent usage in the wild is characterized by permutations and unforeseen situations that deviate significantly from the data used for training.




\section{Proposal: Transformers for HAR}

\subsection{Overview}

Transformers have been shown to have great generalization capability and transfer learning ability \cite{NIPS2017_3f5ee243}. Hence, we propose a Transformer-based architecture, designed explicitly for embedded IMU sensing devices. It is called HART for Human Activity Recognition Transformer. We expect this architecture to better deal with data heterogeneity than both the CNN and DeepConvLSTM models. We also propose an extension called MobileHART which adds layers to better capture temporal dependencies in the sensor data.  


The principle of Transformer-based architectures is to use Multi-Headed Self-Attention (MSA) to dynamically enable the model to give more weight to specific segments of the inputs over other segments. In the context of instance-based (window-based) activity recognition, the entire input window may not contain only relevant information for activity classification. 
Incorporating MSA allows the model to dynamically focus on specific time segments of the signals that best characterize the performing activities (e.g. specific acceleration peaks or significant orientation changes).   


HART uses MSA on the windowed inertial data, segmenting the windows into frames with individual weighted importance. In addition, the complexity of the MSA in HART has been reduced with the use of attention convolutional filters to better fit small devices with limited resources, such as smartphones and smartwatches. However, since transformer-based architectures focus on input-adaptive weighting and global processing, they lack coverage of the spatial inductive biases that CNN-based networks learn. Thus, for this reason, we extended HART to MobileHART, which combines the two architectures and allows the model to encode both local and global information with fewer parameters.

This proposal is inspired by the pioneering work on Transformers in the field of NLP and vision. In particular, the ViT system has produced state-of-the-art results in image processing \cite{khan2021transformers} and has been the building block for many pre-trained models \cite{he2021masked,chen2021empirical,ek4529608comparing}. However, since MSA has quadratic complexity, transformer models can be heavyweights. For this reason, widespread usage of classical transformers in mobile and other low-resourced devices has yet to be adopted. 
Furthermore, to diminish further the computing burden, Wu \emph{et al.} \cite{Wu2020LiteTransformer} have proposed to split the embedding dimension by half, where one half feeds a LightConv block \cite{wu2019pay} while the remaining half feeds an MSA layer. This approach effectively reduces the number of operations performed by the MSA layer. It allows the MSA to focus on capturing the global relationship between inputs instead of local relationships between neighboring segments/patches.

\subsection{HART}

\begin{figure}[htbp]
  \centering
    \includegraphics[width=0.50\linewidth]{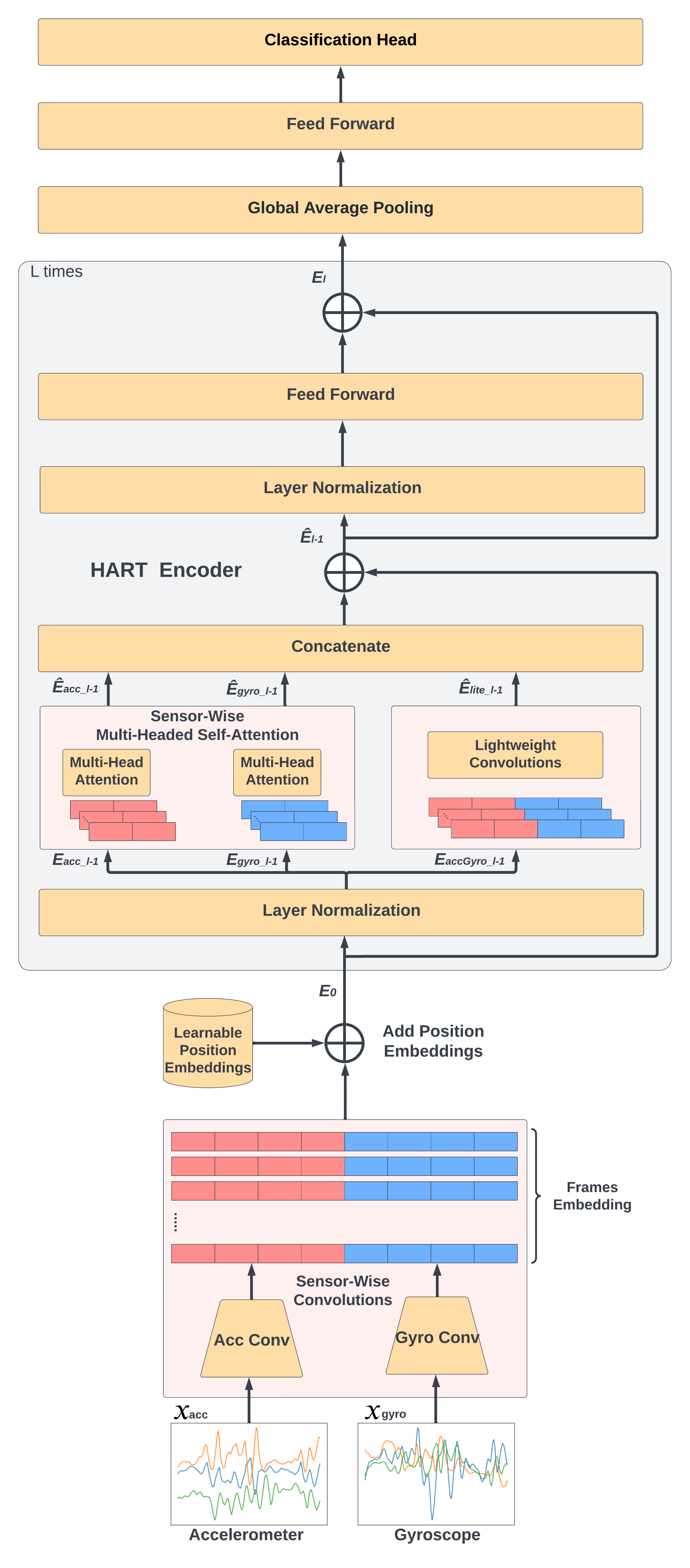}

  \caption{Overview of HART architecture}
  \label{fig:HARToverview}

\end{figure}

The overall architecture of HART, when input data are from accelerometer and gyroscope, is presented in Figure ~\ref{fig:HARToverview}. This architecture highlights the main functional blocks of HART, namely (from bottom to top): 

\begin{itemize}
    \item \textbf{Sensor-Wise Convolutions} for independent  feature extractions per sensors.
    \item  \textbf{Layer Normalization} for flattening local extrema for a smooth fusion between outputs from different sensors.    
    \item \textbf{Sensor-Wise Multi-Headed Self-Attention} for individual attention per the different sensors embeddings.
    \item  \textbf{Lightweight Convolutions} for capturing local attention and embedding sharing between the MSA.
    \item  \textbf{Feed Forward units} for processing the extracted features along with non-linear projections.
    \item  \textbf{Global Average Pooling} for aggregating the differently attended frames together.
    \item  \textbf{Classification Head} for outputting the probability distribution of the inferred activity.

\end{itemize}

\paragraph{Sensors Inputs}

The entire pipeline is now explained. First, let us note that inputs to HART are non-overlapping windows of data $\boldsymbol{x} \in \mathbb{R}^{W,3*S}$ where $W$ is the length of the input window and $S$ is the number of sensors, each having $3$ channels corresponding to the X, Y, and Z axes. For the remainder of the paper, we use a setting where $W = 128$ and $S = 2$ since most datasets contain only accelerometer and gyroscope data. Let us stress, however,  that the architecture has been designed to accommodate more input sources. 


\paragraph{Inputs Embedding $\boldsymbol{E_{0}}$}

At the beginning of the HART process, each sensor data $\boldsymbol{x}$ is partitioned into multiple frames of smaller lengths and projected by specific convolutional layers. Precisely, an accelerometer-oriented convolutional layer processes $\boldsymbol{x_{acc}} \in \mathbb{R}^{W,3}$ and a gyroscope-oriented convolutional layer processes $\boldsymbol{x_{gyro}} \in \mathbb{R}^{W,3}$. Given the embedding dimension hyper-parameter $d$, each convolutional layer has $d/S$ filters and generates non-overlapping frames of size $d/S$ where $d$ has been set to 192. The projection of the accelerometer and gyroscope input data generates N embeddings composed of $\boldsymbol{\hat{x}_{acc}} \in \mathbb{R}^{N,d/S}$ and $\boldsymbol{\hat{x}_{gyro}} \in \mathbb{R}^{N,d/S}$. 
\medskip
\medskip
\medskip

The two outputs of the convolutional layers are concatenated together to form projected embedding vectors. This embedding vectors are then complemented with learnable position embeddings $\boldsymbol{I} \in \mathbb{R}^{N,d}$ to finally have $\boldsymbol{E_{0}} \in \mathbb{R}^{N,d}$. Learnable position embeddings are vectors that encode positions per frame, and $\boldsymbol{E_{0}}$ corresponds to the features extracted from the sensor-wise convolutions that have been encoded with positional information by frames. We do not append a learnable class embedding to $\boldsymbol{\hat{x}}$ and instead favor the use of Global Average Pooling (GAP), which reduces computation cost and parameters with no loss in performances \cite{liu2021swin}. The operations all together can be written as below:
\begin{equation}
 \boldsymbol{E_{0}}= concat(Conv_{acc}(\boldsymbol{x_{acc}}),Conv_{gyro}(\boldsymbol{x_{gyro}})) + \boldsymbol{I} \tag*{}
\end{equation}



$\boldsymbol{E_{0}}$  is then used as the first HART encoder input, which is then processed with layer normalization to standardize the embeddings across different sensors. Since the normalized embeddings were initially concatenations from two different sensors, we can separate back the accelerometer and gyroscope embedding vectors in order to feed each sensor embedding separately to their respective sensor-wise MSA.

\paragraph{HART Encoder: Sensor-wise Attention $\boldsymbol{\hat{E}_{l-1}}$}

HART then partitions and uses the different embeddings as it follows: half of the accelerometer embeddings $\boldsymbol{E_{acc}} \in \mathbb{R}^{N,d/4}$ is processed by an accelerometer-wise Multiheaded Self-Attention ($MSA_{acc\_l}$); 
half of the gyroscope embeddings $\boldsymbol{E_{gyro}} \in \mathbb{R}^{N,d/4}$ is processed by a gyroscope-wise Multiheaded Self-Attention ($MSA_{gyro\_l})$; while the remaining half of both sensors embeddings $\boldsymbol{E_{accGyro}} \in \mathbb{R}^{N,d/2}$ is left concatenated and passed to a $LightConvBlock$ to capture the local relationship and to allow sensor-wise MSA to focus more on global relationships. In addition, as we combine accelerometer and gyroscope embeddings with the $LightConvBlock$, we get an early fusion of the sensor's knowledge. The outline of the process, for $l = 1,2, ... L$, is presented below:
\begin{equation}
\boldsymbol{E_{acc\_l - 1}}, \boldsymbol{E_{gyro\_l - 1}}, \boldsymbol{E_{accGyro\_l-1}} =LN_{l-1}(\boldsymbol{E_{l-1}}) 
\tag*{}
\end{equation}
\begin{equation}
\boldsymbol{\hat{E}_{acc\_l-1}} = MSA_{acc\_l}(\boldsymbol{E_{acc\_l - 1}})   \tag*{}
\end{equation}
\begin{equation}
\boldsymbol{\hat{E}_{gyro\_l-1}} = MSA_{gyro\_l}(\boldsymbol{E_{gyro\_l - 1}})   \tag*{}
\end{equation}
\begin{equation}
\boldsymbol{\hat{E}_{lite\_l-1}} = LightConvBlock_l(\boldsymbol{E_{accGyro\_l - 1}})  \tag*{}
\end{equation}



\begin{figure}[ht]
  \centering
    \includegraphics[width=0.35\linewidth]{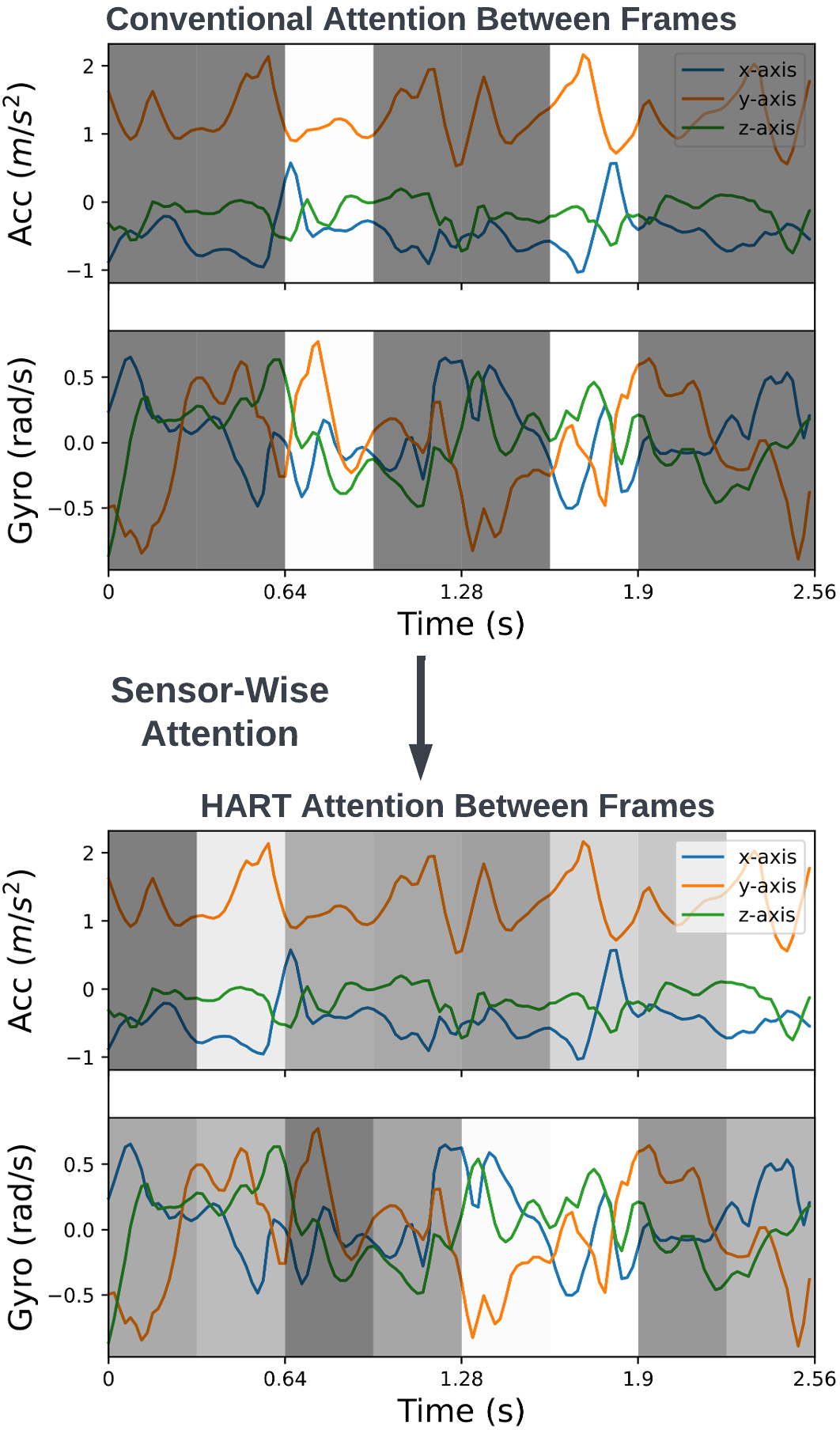}
  \caption{Comparison of attention maps of the accelerometer and gyroscope sensors for a window of the upstairs activity (the darker, the less attention). In HART, MSA can attend to different samples for each sensor.}
  \label{fig:sensorWise}
\end{figure}

Where $\boldsymbol{\hat{E}_{acc\_l}},\boldsymbol{\hat{E}_{gyro\_l}}$ and $\boldsymbol{\hat{E}_{lite\_l}}$ are computed in parallel when supported by the host device. Such architecture allows HART to take advantage of any available on-device GPU to reduce the computation load and inference time. In addition, by adopting the sensor-wise MSA, the model can be separately attentive to different sensor inputs rather than a single attention map over all sensors, as illustrated in Figure~\ref{fig:sensorWise}. This trait allows specialized attention over individual sensors instead of compromised attention between different sensors. This change in architecture is necessary as valuable information is not necessarily at the same time segment between different sensors.

\medskip

The attention mechanism of transformer-based architectures has a complexity that depends on the sequence length, and the projection size $d$ \cite{NIPS2017_3f5ee243}. Thus, splitting the input embedding size to only $1/4$ of $d$ for each sensor-wise MSA can effectively diminish computation needs (without decreasing performances, as experiments will reveal). The original complexity of $O(N^2d )$ of MSA layers would be reduced to $O(2(N^2 d/4))$.

Afterward, $\boldsymbol{\hat{E}_{acc\_l-1}},\boldsymbol{\hat{E}_{gyro\_l-1}}$ and $\boldsymbol{\hat{E}_{lite\_l-1}}$ are concatenated  together, with their original position kept in place, to create $\boldsymbol{\hat{E}_{l-1}}$. 
\begin{displaymath}
\boldsymbol{\hat{E}_{l-1}} = concat(\boldsymbol{\hat{E}_{acc\_l-1}}, \boldsymbol{\hat{E}_{gyro\_l-1}}, \boldsymbol{\hat{E}_{lite\_l-1}}) + \boldsymbol{E_{l-1}}
\end{displaymath}

\paragraph{HART Encoder: output $\boldsymbol{\hat{E}_{l}}$}

The subsequent processes can then be formulated as below:

\begin{displaymath}
\boldsymbol{E_{l}} =  HARTEncoderBlock_{l} = FF_{l-1}(LN{\color{violet}(}\boldsymbol{\hat{E}_{l-1}}{\color{violet})}) + \boldsymbol{\hat{E}_{l-1}}
\end{displaymath}

Since the Feed Forward layer is fed by the concatenation of the two sensor-wise MSA and LightConv outputs, it could be argued that the information about each sensor is lost. However, the residual connection at $\boldsymbol{E_{l}}$ still influences the output values by their sensor-wise position within the vector. Such a property fades as more blocks process the embeddings in the upper blocks. 
The HART encoder block is repeated $L$ times before reaching the classification layers.

\paragraph{Final Classification}

Lastly, a global average pooling layer reduces the dimension of the output of the last HART encoder block through averaging before feeding a feed-forward layer and then finally given to the classification heads.

\subsection{MobileHART}

HART is a sensor-aware and light transformer architecture. However, it does not take full benefit of the convolutional approach that is particularly well suited to learn temporal inductive biases \cite{47094} as reported in other HAR studies using Transformer architectures \cite{IFConvTransformer2022,Kim2022}. Inspired by MobileViT \cite{mehta2022mobilevit}, we improved the capability of the HART encoder with lightweight CNN layers at the input and output to devise the MobileHART block. These CNNs make it possible to compute local temporal relations while maintaining the temporal coherence of the output data through unfolding and folding operations. 

\begin{figure*}[ht]
  \centering
      \includegraphics[width=\linewidth]{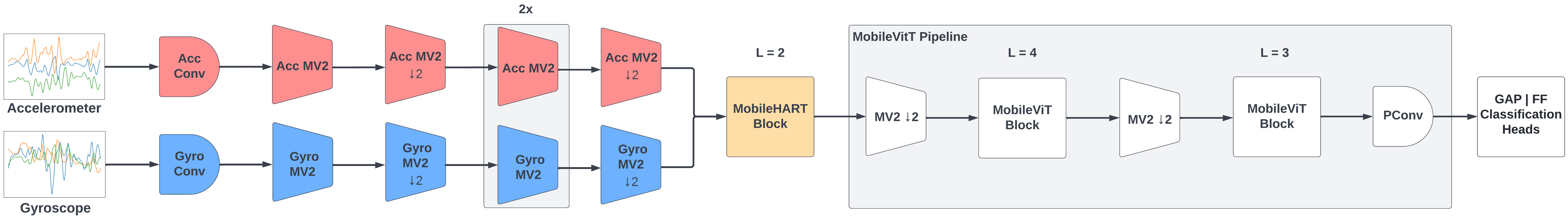}

  \caption{Overview of MobileHART}
  \label{fig:mobileHART}
\end{figure*}

As illustrated in Figure~\ref{fig:mobileHART}, before feeding the MobileHART block, the input representation is improved using sensor-wise convolutional layers and sensor-wise inverted residual blocks (AccMV2, GyroMV2) to process sensor inputs separately. The MV2 blocks are composed of pointwise and depth-wise convolutional layers \cite{howard2017mobilenets}, that are used for down-sampling and embeddings. The outputs of the two sensors' convolutional processes are concatenated to form the inputs to the MobileHART block. This block uses sensor-wise pointwise convolutional layers ($PConv_{SW}$) and standard convolutional layers ($SConv_{SW}$) to capture both local and global information. We note that the sensor-wise operation is performed only up to the input of the MobileHART block, where afterward, the features from different sensors are fused. The entire process described here can be expressed as below:

\begin{equation}
  AOut(\boldsymbol{x}) = HARTEncoderBlocks_L(PConv_{SW}{\color{orange}(}SConv_{SW}{\color{violet}(}\boldsymbol{x}{\color{violet})}{\color{orange})}) \tag*{}
\end{equation}
\begin{equation}
  MobileHARTBlock(\boldsymbol{x}) = SConv_{2}(Concat{\color{teal}(}PConv_{2}{\color{orange}(}AOut{\color{violet}(}\boldsymbol{x}{\color{violet})}{\color{orange})},\boldsymbol{x}{\color{teal})}) \tag*{}
\end{equation}

where $\boldsymbol{x} \in \mathbb{R}^{N,d}$ is a feature embedding that has been processed by multiple layers of accelerometer and gyroscope MV2 blocks. Afterward, the remaining operations follow the pipeline of the MobileViT architecture process that we adapted to fit one-dimensional time-series data. This model results in a longer processing chain but with a greater ability to compute local (via CNN) and global (via Transformers) representation. This ability is empirically exhibited in Section~\ref{sec:newEvaluation}.

\section{Experiments}\label{sec:hyper_parameters}

\subsection{Configuration}
 
 We performed 200 epochs of class-weighted training to deal with high-class imbalance for all our experiment settings while using a mini-batch size of 32. We then used a label smoothing rate of $0.1$ to prevent overfitting. Across all learning instances, we employed Adam \cite{kingma2014adam} as the learning optimizer with a learning rate of 0.0005. Based on other studies on low-resource data \cite{araabi-monz-2020-optimizing}, we settled on a dropout rate of $0.3$ after most feed-forward layers. We note that our dropout rate is much higher than the usual choices with respect to other works on transformer-based models; usually, a rate of $0.1$ - $0.2$ \cite{dosovitskiy2021an,Wu2020LiteTransformer,caron2021emerging}. This choice is because, as opposed to the vision or language domain, the size of the datasets in the HAR domain is significantly smaller, which makes the models more prone to overfitting. As an extension to counter overfitting, we also applied DropPath with regards to the stochastic depth rule before every residual connection \cite{larsson2017fractalnet,huang2016deep}. We found that swish activation \cite{https://doi.org/10.48550/arxiv.1710.05941}, in line with other studies \cite{mehta2022mobilevit}, gave better results than the classical GELU activation \cite{https://doi.org/10.48550/arxiv.1606.08415} used in transformer-based models.

 

 \subsection{Performances with known situations}

\begin{table*}
  \caption{F-score with known situations on the UCI, MotionSense, HHAR, RealWorld, and SHL test sets. \textbf{Bold} indicates best results while \underline{underline} indicates second-best results.}
  \label{tab:fourResult}
  \centering
  \resizebox{\columnwidth}{!}{
  \begin{tabular}{ccccccc}
    Architecture & UCI & MotionSense & HHAR & RealWorld &SHL & Combined\\
    \hline
     CNN \cite{Ignatov2018RealtimeHA} &  94.53 & 97.72   & 96.91 & 92.62    &  75.33 & 80.98\\
     DeepConvLSTM \cite{ordonez2016deep} & \underline{95.94} & \underline{98.17} & \underline{97.94} & 94.39  & \underline{ 80.31 } & 73.46 \\
    \hline
     Conventional Transformer \cite{dosovitskiy2021an} & 93.66 & 97.56 & 96.05 &  93.80   &  77.41 & 84.21\\
      HART [Ours] & 94.49 &   98.20 & 97.36 & \underline{94.88} & 79.49 & \underline{85.61} \\
     MobileHART [Ours]& \textbf{97.20} & \textbf{98.45} & \textbf{98.19} & \textbf{95.22} &  \textbf{81.36} & \textbf{86.74}\\
    \hline
  \end{tabular}
  }
\end{table*}


Table~\ref{tab:fourResult} presents the F-score obtained by the different model architectures on the five datasets and on the combination of the five. Regarding the latter, let us recall that we used class-weighted training to deal with the high-class imbalance. Apart from CNN and DeepConvLSTM, we also included a conventional Transformer (CT) encoder \cite{NIPS2017_3f5ee243}. We note here the conventional Transformer and HART were implemented from the same initial base code \cite{dosovitskiy2021an} and share the same hyper-parameters for fairness in evaluation (an embedding size of 192, 6 encoder transformer blocks, three attention heads per block, Swish activation, and a drop-out of 0.3). Comparison with recent Transformer based HAR models \cite{IFConvTransformer2022,Kim2022} was not directly possible since most papers do not publicly release their code and there is no consensual benchmark in HAR for mobile phones. This is why, we have made the data partition and our code publicly available\footnote{https://github.com/getalp/Lightweight-Transformer-Models-For-HAR-on-Mobile-Devices} to avoid such lack of reproducibility in the domain.

Overall, we can see different behaviors according to the size of datasets size and the degree of heterogeneity. The CNN approach is very competitive on homogeneous datasets, such as UCI but clearly underperforms on larger and more challenging datasets. DeepConvLSTM, performs even better than CNN in the majority of the cases. However, on the large combined dataset, we see that the performance of the model significantly drops, performing the worst among the tested models.

The Conventional Transformer and HART have an inverse behavior showing superiority over the CNN and DeepConvLSTM when datasets become bigger. More so HART was able to outperform the conventional transformer in all cases. MobileHART models perform best on all datasets. In particular, MobileHART achieves the best performance on all five datasets. HART performs slightly worse than MobileHART but shows consistent improvement over the Conventional Transformer. Overall, it is interesting to note that the score gap between CNN and Transformers increases with the size of the training dataset, with the Combined case showing the largest gap. This confirms the ability of the Transformer models to take better advantage of a larger training set than more traditional architectures. 


Finally, the slightly better performance of HART than MobileHart on the combined dataset can be due to the higher embedding space size of HART (196 vs. 144). The combined dataset represents more complex functions (situations) than individual dataset training and the model would need a greater capacity to be able to represent this complexity. This behavior would be in line with other studies, where transformer-based models trained from large sets of diverse datasets require a larger capacity~\cite{https://doi.org/10.48550/arxiv.2106.10270}.

\subsection{Performances with unseen situations}


\begin{table*}
  \caption{F-score with unseen body positions on RealWorld  }
  \label{tab:hetereogenityTableRW}
  \centering
  \resizebox{\columnwidth}{!}{
  \begin{tabular}{c|ccccccc|c}
    Architecture & Chest & Forearm & Head & Shin & Thigh & Upperarm & Waist & Average $\pm$ STD \\
    \hline
    CNN &  \underline{67.09} & \underline{37.19} &\underline{47.80}& \underline{45.50} & \underline{51.76} & \underline{56.83} & \underline{45.78} & \underline{50.27} $\pm$ 8.85\\
    DeepConvLSTM &  61.28 & 31.28 &41.89& 43.97 & 50.20 & 53.80 & 37.63 &47.07 $\pm$ 10.41 \\

    \hline
    HART &  54.15 & 26.32 &42.71& 38.98 & 45.92 & 48.46 & 39.35 &42.17 $\pm$  8.20 \\
    MobileHART & \textbf{74.58} & \textbf{49.76} &  \textbf{61.48} & \textbf{47.43} & \textbf{54.75} & \textbf{69.34} & \textbf{50.62} & \textbf{58.28 $\pm$ 9.70}\\
    \hline
  \end{tabular}
  }
\end{table*}

\begin{table*}
  \caption{F-score with unseen devices on HHAR }
  \label{tab:hetereogenityTableHHAR}
  \centering
  \resizebox{\columnwidth}{!}{

  \begin{tabular}{c|cccccc|c}
    Architecture  & S.Gear & LG.Watch & Nexus4 & S3 & S3 Mini & S+ & Average $\pm$ STD \\
    \hline
    CNN &   \underline{65.58}  & 59.88 & \underline{96.33}& 85.70 & 91.46 & 78.28 & 79.53  $\pm$ 13.19\\
    DeepConvLSTM & 60.47 & 42.05 &91.46& 90.71 & 88.84 & 83.25 &  76.13 $\pm$  20.32\\

    \hline
    HART  &  63.62 & 60.62 &95.88& \textbf{93.51} & \underline{93.83} & \underline{92.30} & \underline{83.29} $\pm$ 15.03\\
    MobileHART & \textbf{81.47} & \textbf{81.37} &  \textbf{98.32} & \underline{91.41} & \textbf{97.25} & \textbf{94.72} & \textbf{90.75 $\pm$ 6.95}\\
    \hline
  \end{tabular}
  }
\end{table*}

Results with unseen positions and devices for each model are respectively presented in Tables \ref{tab:hetereogenityTableRW} and \ref{tab:hetereogenityTableHHAR}. Here the evaluation method presented in Section~\ref{sec:evaluation} to demonstrate the heterogeneity problem has been reapplied, and the F-score of the four considered architectures are shown. 

Table~\ref{tab:hetereogenityTableRW} shows a severe performance degradation for all the models. MobileHART clearly outperforms all other architectures in all body positions. The second best is the CNN model, with more than 13\% absolute differences with MobileHART. It seems that models which include several CNN layers in their process are better able to transfer learning to unseen positions.  

Table~\ref{tab:hetereogenityTableHHAR} shows the results with unseen devices. Not having the evaluated device in the training set appears to create a lesser detrimental effect than for the body position experiment. HART and CNN both obtained close scores. The DeepConvLSTM performs the worst with performance as low as 42.05\% on the LG.Watch. MobileHART again, except for the Samsung S3 device,  outperforms all other models with a score as high as 98.32\% on the Nexus 4 device. This model's F-score is also much more stable than other architectures with a standard deviation of only 6.95\% across all devices. As previously emphasized, the Samsung Gear and LG smartwatches were positioned on the wrist, while the phones used for training were conveyed in a hip-mounted pouch. Hence, for these cases, the models must address both system and statistical heterogeneity. This is where MobileHART exhibits a major improvement in performance.

Our results here show that the mismatch of body positions (statistical heterogeneity) impacts the model's performance more than the one of devices (system heterogeneity). The RealWorld dataset best illustrates this impact where the best average score across all body positions was at 58.28\% achieved by MobileHART. The CNN approach was about 8\% behind in terms of performance. The gap between the MobileHART and the CNN was further widened on the HHAR dataset, where the average performance on different devices was about 11\% in differences.

The study also shows that the Transformer models alone are insufficient for very challenging tasks such as the mismatch of body position (Table~\ref{tab:hetereogenityTableRW}). Indeed, HART exhibits 42.17\%  on the averaged scores. This finding is in line with the literature \cite{mehta2022mobilevit}, where both temporal capabilities provided by CNNs and the input-adaptive weighting with global processing provided by the attention-based model are necessary to reach a better generalization.


\subsection{Varying Sensors}\label{multisensor}
Here we explore the capabilities of HART to accommodate different input sources. Specifically, we investigate the inclusion of a magnetometer to measure magnetic field intensities, aiming to assess whether HART can effectively scale with multiple sensors. It is important to note that the embedding dimension, denoted as $d$, remains constant even with the inclusion of more sensors. For instance, when 3 different sensors are present and the embedding dimension is of size 192, the embedding shapes are distributed as follows: 96 for LightConv ($d/2$), and 32 for each of the three sensors in MSA ($d/6$). If we were to include four sensors, the embedding size would be further reduced to 24 for each sensor ($d/8$).

The findings are presented in Table~\ref{tab:sensorStudy}, which shows that HART can adapt with an additional sensor where we see a 1.00\% gain when the magnetometer is added on top of the Accelerometer and Gyroscope. Additionally, we trained different combinations of sensors where we used accelerometer and magnetometer sensors, which showed a slight improvement in results (0.07\% gain over accelerometer and gyroscope). In HAR on mobile devices, using fewer sensors allows developers more freedom in terms of cost and product choices. The results here thus hint that only two sensors may be sufficient for HAR, as using the magnetometer and other less common sensors adds little performance gains \cite{overviewHar}.

\begin{table}[h]
    \centering

  \caption{F-Score of HART model on MotionSense with different sensors} 
  \label{tab:sensorStudy}
  \begin{tabular}{lr}
    \toprule
    Sensors & F-Score \\
    \midrule
    Accelerometer, Gyroscope & 98.13 \\
    Accelerometer, Magnetometer & 98.21 \\
    Accelerometer, Gyroscope, Magnetometer & 99.00 \\
  \bottomrule
\end{tabular}
\end{table}

\subsection{On-Device Performances}

  \begin{table*}
  \caption{On-Device inference time per inference and model footprint computed over 1000 inferences}
  \label{tab:inferenceTable}
  \centering
  \resizebox{\columnwidth}{!}{
  \begin{tabular}{cc|cccc}
    \hline
    Architecture & Avg Inference Time (ms)   & Memory (MB) & Model Size (MB) & Parameters & FLOPS\\
        \hline

    CNN \cite{Ignatov2018RealtimeHA}  & \textbf{2,14 $\pm$ 0.45} & 51.05 & 25.80 & 6,448,714 & 17,725,476\\

    DeepConvLSTM \cite{9065078} &40.90 $\pm$ 2.75 & \textbf{10.16 } & \textbf{1.85}& \textbf{457,288} & 93,496,654 \\

     \hline

    Conventional Transformer \cite{dosovitskiy2021an} & 8.21 $\pm$ 1.51 &  32.07& 15.22 & 3,783,238 & \underline{17,069,949}\\
    HART [Ours] &  \underline{5.38 $\pm$ 1.10} &  \underline{12.74} & \underline{5.91}&\underline{1,445,918} & \textbf{15,212,636}\\

    MobileHART [Ours]& 13.49 $\pm$ 1.47 & 20.77 & 10.35&2,542,942 &19,809,292\\

  \hline
\end{tabular}
}
\end{table*}
 We ported all models in our experiments, using TensorFlow lite, to a 2017 Google Pixel~2 with a Qualcomm~MSM8998 Snapdragon~835 chipset (octa-core CPU with 4x2.35~GHz~Kryo \& 4x1.9~GHz~Kryo) and 4 GB of RAM. Afterward, we use TensorFlow's benchmark tools to measure the average inference time along with its standard deviation, the memory footprint, and the model size for over 1000 inferences using 4 threads with the different architectures\footnote{\url{https://www.tensorflow.org/lite/performance/measurement}}. For technical reasons, all models were inferred using the on-device CPU.

The results are presented in Table~\ref{tab:inferenceTable}. The inference time of the models is between 1 ms and 41 ms for an input window of 2.56 seconds. The transformer-based model with the longest inference time in our study, MobileHART, can classify more than 76 samples per second, which is well above the real-time criteria of 28 samples per second set by another study \cite{Ignatov2018RealtimeHA}. This performance shows that all the models studied in this paper can achieve real-time activity classification. 

We first see that CNN with an average inference time of 2.14 ms. Transformer-based architectures require a longer inference time, as shown with MobileHART, which has an average inference time of 13.49 ms. This is mainly due to
its relatively high number of stacked layers. On the other hand, we are able to observe a relatively much lower inference time with HART. The incorporated lightweight components of HART allowed the model to outperform the conventional transformer (5.38 ms against 8.21 ms). Lastly, here, we stress that implementations of HART are developed to take advantage of parallel computations. If the workload is done sequentially or with little parallelization given only a few computational threads, we may see an additional inference duration increase, as observed with MobileHART. The slowest model, however, is the DeepConvLSTM with an average inference time of 40.9 ms. This is likely due to the LSTM component that requires higher sequential processing time.

In terms of memory and model size during inference, the number of parameters plays a crucial role. Among the evaluated models, DeepConvLSTM stands out as the lightest, with a memory footprint of 10.16 MB and a model size of 1.85 MB. In contrast, the CNN model, with approximately 6 million parameters and 17 million FLOPS, occupies a total of 51.05 MB of RAM for inference and requires 25.80 MB of storage space. Comparatively, MobileHART offers a more compact solution, consuming less than 2 times the resources in terms of memory (20.77 MB), model size (10.35 MB), and parameters (around 2.5 million), except for FLOPS, where MobileHART performs around 11\% more operations. Finally, It is worth mentioning that HART, despite requiring approximately 50\% fewer parameters than the Conventional Transformer (1,445,918 versus 3,783,238 parameters), consistently exhibits superior classification performance. This significant parameter reduction also results in a notable decrease in the model's memory and storage footprint.

\section{Analysis and discussion} \label{sec:newEvaluation}

To interpret the results obtained we made a visual T-SNE analysis \cite{van2008visualizing} of the CNN, HART, and MobileHART models trained on the RealWorld and HHAR datasets, by projecting the instances of the test set on the last layer (before activation) of the neural network. The objective is to assess qualitatively the embeddings by checking whether different classes are well separated from each other and if the values of the same class remain close despite different positions or different devices. 

Precisely, we looked into the representations of the three mentioned models on the 7 body positions of the RealWorld dataset to estimate the model's robustness against \emph{statistical heterogeneity} and on the 6 devices of the HHAR dataset to estimate the robustness against \emph{system heterogeneity}.

\begin{figure*}[!htb]
\minipage{0.30\textwidth}
  \includegraphics[width=\linewidth]{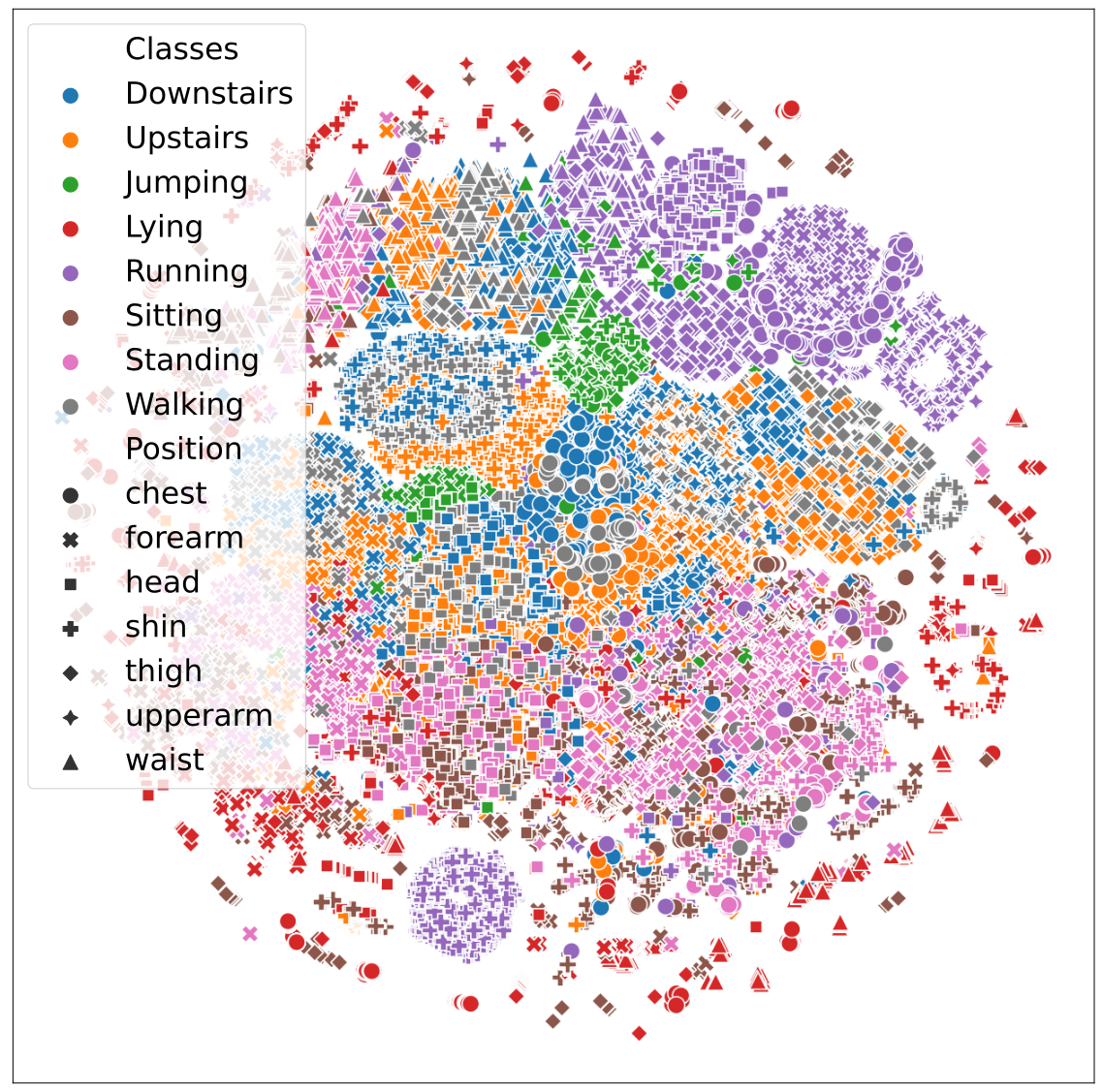} 
  \caption{CNN RW}\label{fig:CNNRW}
\endminipage\hfill
\minipage{0.30\textwidth}
  \includegraphics[width=\linewidth]{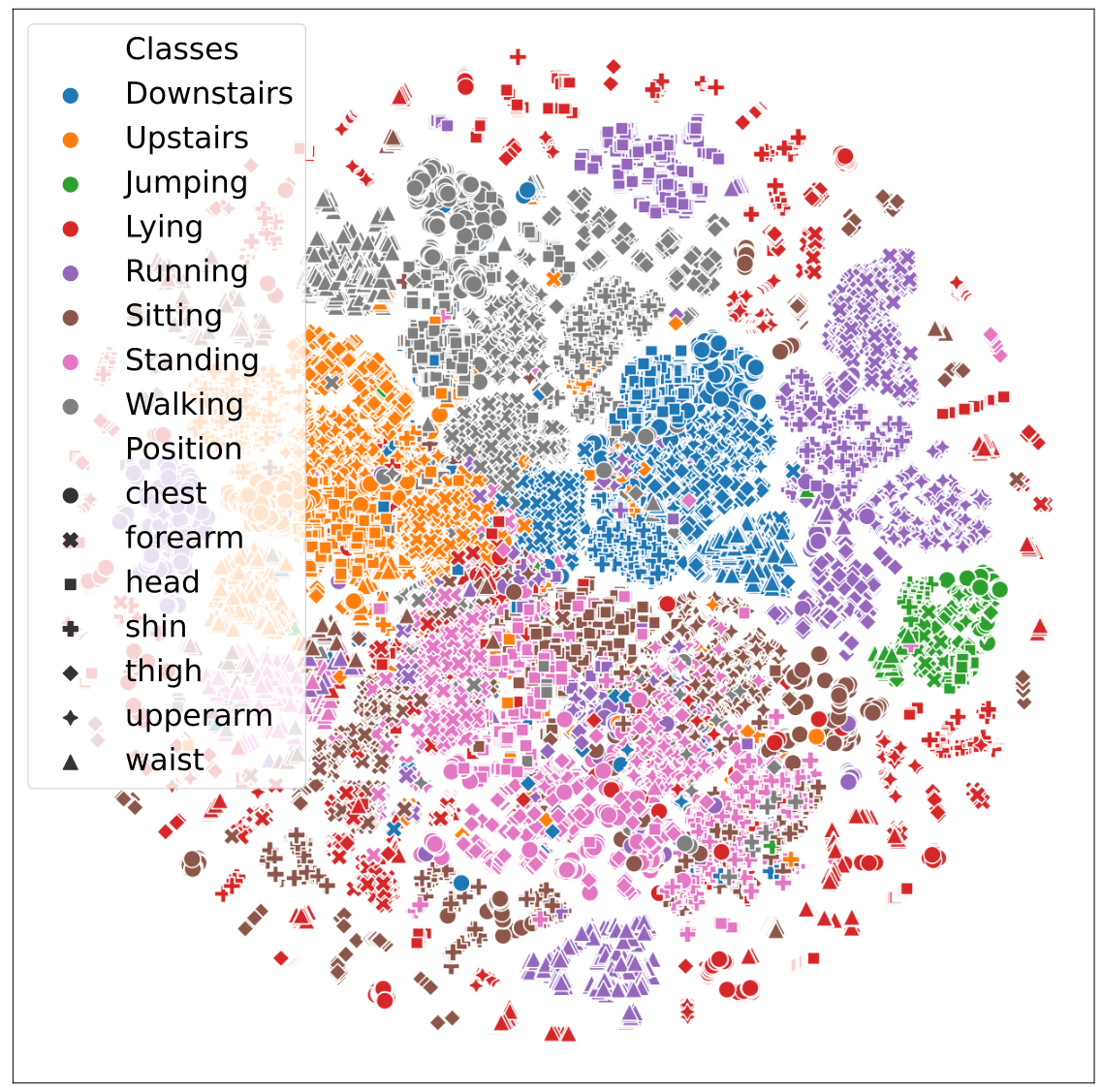} 
  \caption{HART RW}\label{fig:HARTRW}
\endminipage\hfill
\minipage{0.30\textwidth}
  \includegraphics[width=\linewidth]{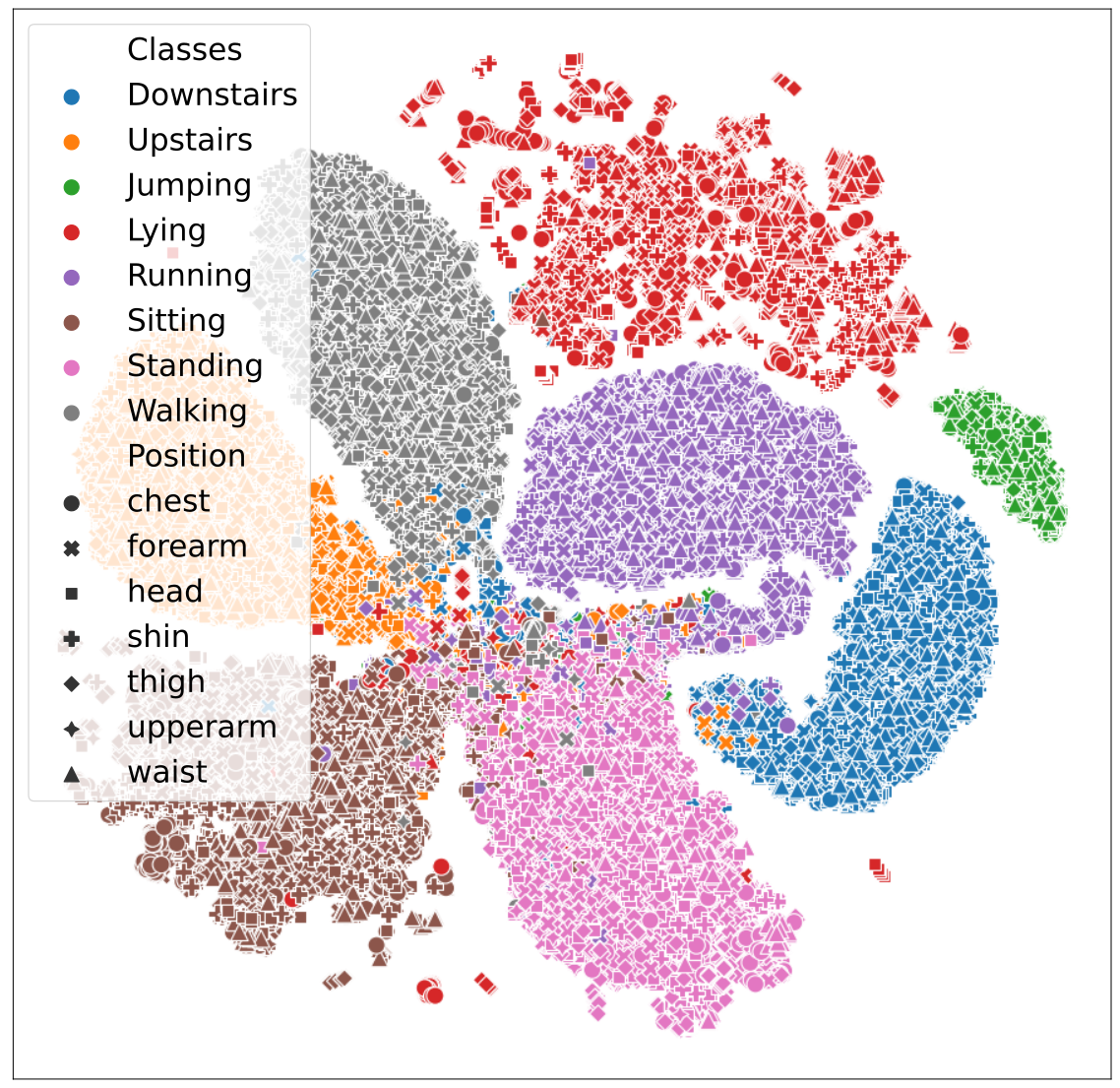} 
  \caption{MobileHART RW}\label{fig:MOBILEHARTRW}
\endminipage\hfill
\end{figure*}

\begin{figure*}[!htb]
\minipage{0.29\textwidth}
  \includegraphics[width=\linewidth]{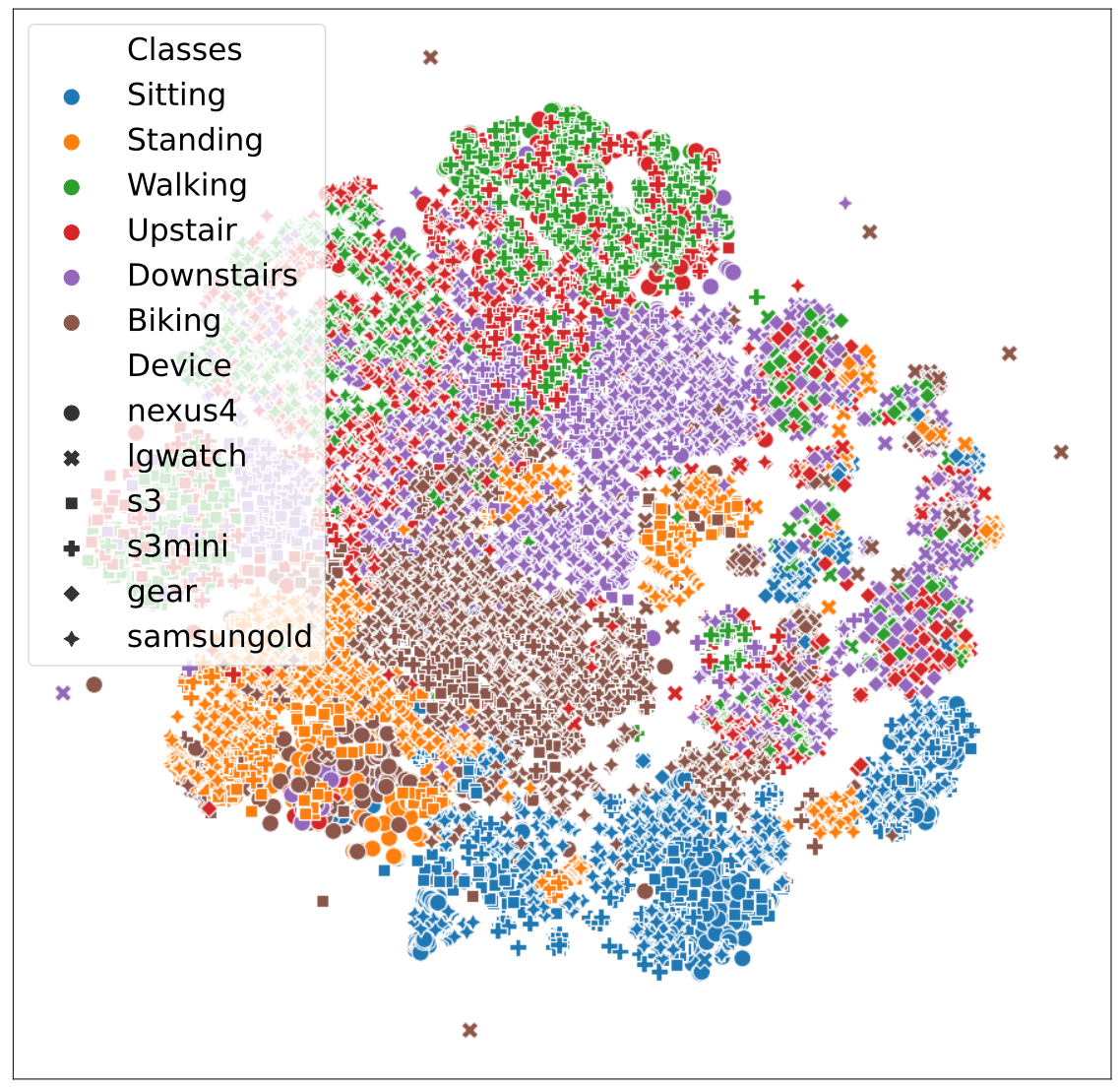} 
  \caption{CNN HHAR}\label{fig:CNNHHAR}
\endminipage\hfill
\minipage{0.29\textwidth}
  \includegraphics[width=\linewidth]{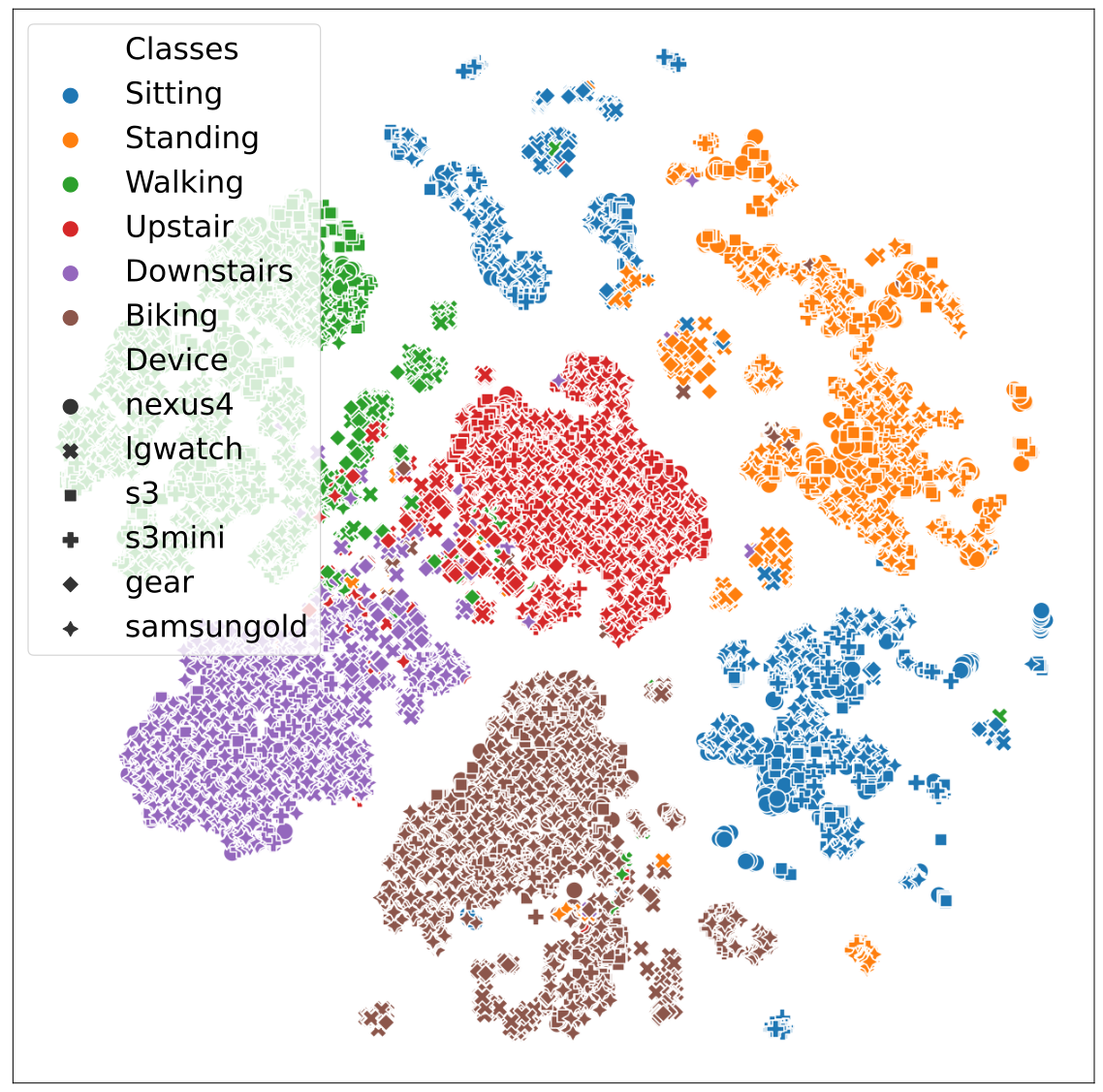} 
  \caption{HART HHAR}\label{fig:HARTHHAR}
\endminipage\hfill
\minipage{0.32\textwidth}
  \includegraphics[width=\linewidth]{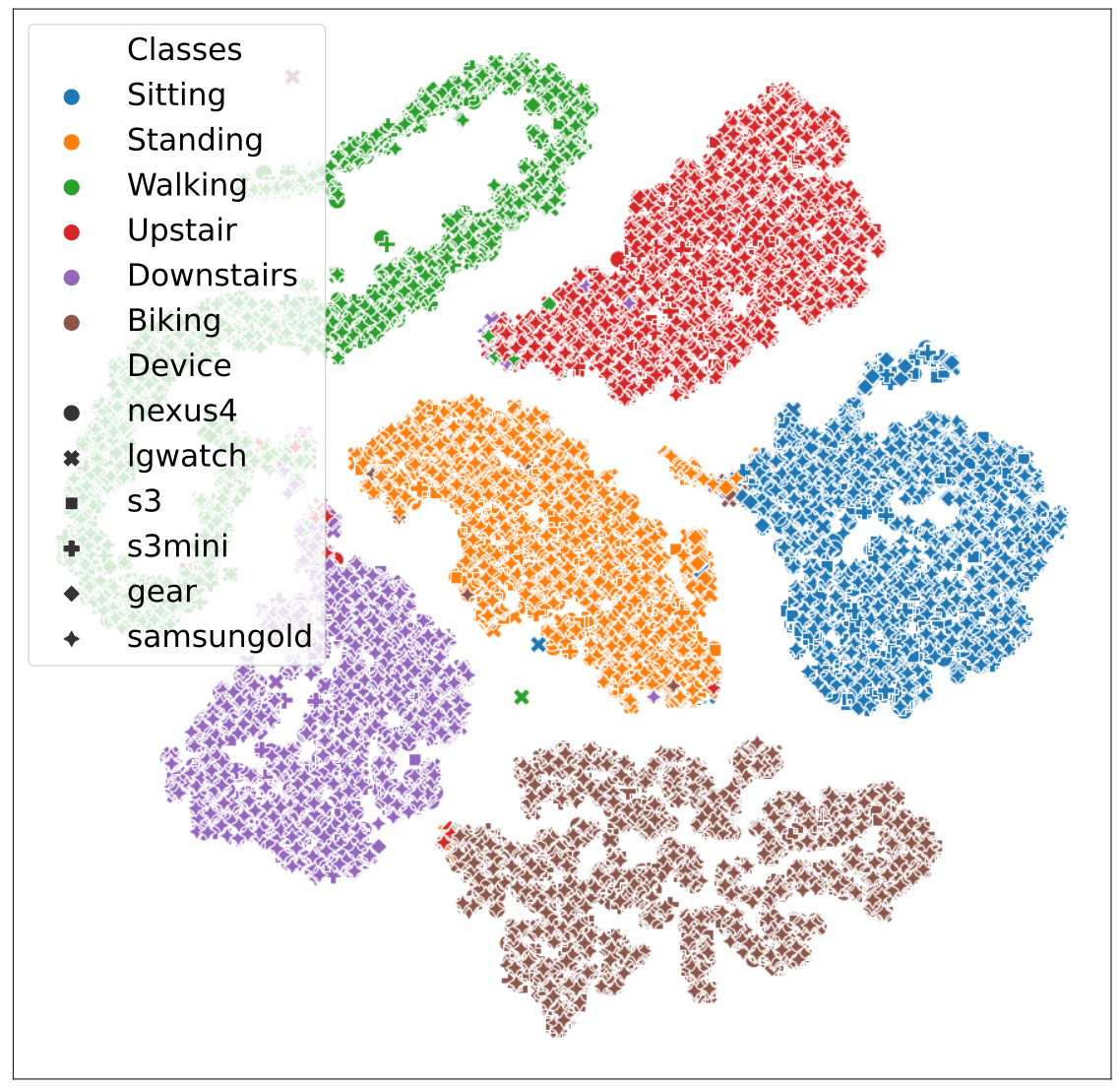} 
  \caption{MobileHART HHAR}\label{fig:MOBILEHARTHHAR}
\endminipage\hfill
\end{figure*}

Regarding the CNN model, results of the T-SNE analysis are given by Figures~\ref {fig:CNNRW} and \ref{fig:CNNHHAR}. Each projected instance is represented by a color (its activity) and a shape (its body position or devices). It can be seen this model manages to separate the activities but in an imperfect way. On the one hand, movement-related activities and static activities are clearly separated. On the other hand, distinctions are more difficult to make within these two clusters. For instance, the Downstairs and Upstairs motion-based activities and the Standing and Sitting fixed activities are mixed together. In addition, we also see that the change of position is difficult to manage. This trait is highlighted by the Running activity, where the Upperarm position forms a separate cluster far away from all other positions. Similarly, the Lying activity instances were not clustered but distributed around all other activities. 

The representations on the HHAR dataset, on Figures \ref{fig:CNNHHAR}, show better separation between classes, except for the Walking and Upstairs activities which are often mixed. Furthermore, the activities are not individually clustered, and the representation of devices is segregated.

 For the HART model, we can easily observe a much better separation of classes than the CNN model (cf. Fig.~\ref{fig:HARTRW} and \ref{fig:HARTHHAR}). For instance, Downstairs and Upstairs activities are now well separated and clustered on the RW dataset. However, fixed activities like Sitting and Standing are still intertwined. Also, the Lying activity remains at the edge of all activities without its own cluster. Conversely, the embeddings of HART on the HHAR dataset significantly improve the clustering of activities, except for the Sitting activity, where the representations are broken into two well-separated clusters.

Finally, for the MobileHART model (cf. Fig. ~\ref{fig:MOBILEHARTRW} and \ref{fig:MOBILEHARTHHAR}), we can see a clear improvement since the activities are now clearly separated in both the RW and HHAR datasets. The representations from different positions or devices are respectively well clustered together with the activities. Specifically, on the RW dataset, the Lying activity that was spiraling around other activities in the previous two model representations is now well-formed into a single cluster. On the HHAR dataset, the Sitting activity is now grouped into a single cluster.

It is easy to understand, intuitively, that a good separation of the activities embeddings facilitate the work of classification \cite{caron2021emerging}. Indeed, the separation power of the learned embeddings of MobileHART is correlated to its superior classification performances.  
Also, good embeddings should also allow an easier adaptation to new situations such as model fine-tuning on a new device or position as demonstrated in other domains \cite{khan2021transformers}. 

\section{Conclusion \& Future Works}



In this paper, we argue that despite the impressive accuracy reported in the literature of ML models in the wearable Human Activity Recognition (HAR) domain, there is a strong need for improved model robustness.  Indeed, most models trained on a specific HAR dataset, cannot withstand data heterogeneity that is particularly acute in pervasive applications. This heterogeneity arises in pervasive application due to statistical heterogeneity (differences in usage and environment) and system heterogeneity (differences in system traits). In this study, we demonstrated that traditional deep neural networks such as CNN and CNN-LSTM which have been the state-of-the-art in the HAR domain \cite{gu2021survey} can dramatically suffer from statistical heterogeneity and system heterogeneity. For instance, a CNN model showing a 91\% F-score on the RealWorld dataset cannot exceed 67\% at best (37\% in the worst case) on a smartphone position it has not been trained with.

We argue that this problem is not only a problem of data mismatch but also a failure in learning good representation. To tackle this latter problem, we introduced Human Activity Recognition Transformer (HART), a novel Transformer-based model in the HAR on mobile devices domain. The model extends ViT by using lightweight convolutions and sensor-wise architectures to fit the multi-sensor input and computing resources constraint of the HAR domain. HART was then expanded with convolutional filters to provide it with a greater ability to learn local (via CNN) together with global (via Transformers) representation. This architecture has been named MobileHART. The experiment done on five publicly available datasets and MobileHART, outperforms previous state-of-the-art CNN and CNN-LSTM models on all datasets. In addition, the baseline HART also was able to be very competitive when given a larger amount of data. HART and MobileHART exhibit a stronger improvement (up to 8\% absolute F-score difference) with large and more diverse datasets showing that such models are much more adequate to capture diverse situations and leverage a large amount of data. 

Regarding robustness, MobileHART is the least impacted by statistical heterogeneity (here body position of the device) with an 8\% average F-score difference against CNN. For system heterogeneity (difference of devices), MobileHART shows great stability with a 10\% average F-score difference with CNN and the smallest standard deviation. Hence, MobileHART showed the best robustness in handling heterogeneity whether unseen mobile placement positions or device brands. We argue that this is due to its better ability to learn mode generic representation as the qualitative space embedding projection has revealed. This property makes it relevant for on-device adaptation to clients as well as to clustering-based techniques such as active learning \cite{Presotto2022}. Further methods to address such heterogeneity also include domain adaptation and supervised and unsupervised pre-training \cite{10207558}. To ease the extension of this work and to encourage reproducibility, the code, and data partitions have been made publicly available at the following link \footnote{https://github.com/getalp/Lightweight-Transformer-Models-For-HAR-on-Mobile-Devices}.  



This study has demonstrated that thanks to efficient light attention-based mechanisms and convolutional filters, HART and MobileHART can indeed run on mobile devices, exhibiting superior performances with far lower memory footprint than a 2-layer CNN model. Such a model would thus be adapted to on-device learning in the form of federated learning \cite{mcmahan2017communication} with HART. Recent studies have shown that it is possible to integrate transformers into a collaborative learning ecosystem \cite{park2021federated}. Such an approach would give the learning access to a far greater training data distribution than classical central learning. Since HART and MobileHART can handle large training data, they are promising robust models to address the deployment of HAR ML models in real-life heterogeneous environments \cite{sannara2021federated}.

\section*{Acknowledgments}
This work has been partially funded by Naval Group, and by MIAI@Grenoble Alpes (ANR-19-P3IA-0003). This work was also granted access to the HPC resources of IDRIS under the allocation 2022-AD011013233 made by GENCI.

\section*{Data Availability}
All the datasets used during the study are described in Section~\ref{sec:datasets} and are available through the following links: 
\begin{itemize}
    \item UCI: \url{https://archive.ics.uci.edu/ml/datasets/human+activity+recognition+using+smartphones}
    \item MotionSense: \url{https://github.com/mmalekzadeh/motion-sense/tree/master/data}
    \item HHAR: \url{http://archive.ics.uci.edu/ml/datasets/Heterogeneity+Activity+Recognition}
    \item RealWorld (RW): \url{https://sensor.informatik.uni-mannheim.de/#dataset_dailylog}
    \item SHL: \url{http://www.shl-dataset.org/download/}
\end{itemize}

\section*{Conflict of Interest}
The authors have no relevant financial or non-financial interests to disclose.

\bibliographystyle{plain}
\bibliography{sn-bibliography} 
\end{document}